\documentclass{extarticle}
\usepackage{float}
\usepackage{graphicx} % Required for inserting images
\usepackage{tabularray}
\usepackage{adjustbox}
\usepackage{xcolor}
\usepackage{graphicx} % Required for inserting images
\usepackage{subcaption}

\title{From Linear to Spline-Based Classification: Developing and Enhancing SMPA for Noisy Non-Linear Datasets}
\author{ Vatsal Srivastava \\ 
        Department of Computational Intelligence\\
       School of Computing\\
       SRM Institute of Science and Technology\\
       Chennai, Tamil Nadu, India}
\date{March 2025}

\begin{document}

\maketitle

\begin{abstract}
    Building upon the concepts and mechanisms used for the development in Moving Points Algorithm, we will now explore how non linear decision boundaries can be developed for classification tasks. First we will look at the classification performance of MPA and some minor developments in the original algorithm. We then discuss the concepts behind using cubic splines for classification with a similar learning mechanism and finally analyze training results on synthetic datasets with known properties. 
\end{abstract}

\section{Introduction}
The development of Moving Points Algorithm \cite{mpa} opened a new avenue to machine learning with the introduction of heuristic learning mechanisms rather than optimization of loss or cost functions. Initially, the learning algorithm was unstable and did not perform well on performance metrics. The linear algorithm was tested on 10 standard datasets\cite{diabetes}\cite{iris}\cite{penguin}\cite{mammographic}\cite{sonar}\cite{wisconsin}\cite{ionosphere}\cite{parkinson}\cite{wdbc}\cite{wpbc}. Afterwards, different feature selection techniques were used to test the algorithm on reduced features and benchmark against existing industry standard algorithm. 

Considering the algorithm's inability to create non linear decision boundaries, changes were made to how a decision boundary is created and cubic splines were incorporated to model non linear decision boundaries by increasing the number of control points.

\section{Complete Evaluation of MPA}

The evaluation of Moving Points Algorithm on different 10 different datasets with varying number of features and class distributions. Popular datasets from UCI machine learning repository were used for evaluation across 4 metrics. Feature selection and hyper-parameter tuning were also given importance in evaluation. The current stage of evaluation only focused on processing of numerical data available in csv formats for binary classification only. 

\section{Summarized findings}
The evaluation revealed that the proposed algorithm has reduced accuracy in datasets with high dimensionality ($>$60 features). The proposed algorithm did not achieve the highest CV accuracy in any of the datasets tested. 

\begin{table}[h!]
    \centering
    \small
    \resizebox{\textwidth}{!}{ 
        \begin{tblr}{
            colspec={c|c|c|c|c|c|c},
            row{odd} = {gray!15}, % Light gray for odd rows
            row{even} = {white}   % White for even rows
        }
        Dataset               & KNN   & Perceptron & Random Forest & Decision Tree & SVM   & MPA    \\
        \hline
        Diabetes              & 0.584 & 0.637      & 0.667         & 0.587         & 0.632 & 0.649  \\
        Iris~                 & 0.977 & 0.974      & 0.963         & 0.963         & 0.977 & 0.9411 \\
        Penguin               & 0.995 & 1.0        & 0.987         & 0.975         & 1.0   & 0.990  \\
        Mammoraphic           & 0.807 & 0.807      & 0.830         & 0.809         & 0.833 & 0.819  \\
        SONAR                 & 0.872 & 0.785      & 0.882         & 0.777         & 0.787 & 0.729  \\
        Wisconsin(9 features) & 0.963 & 0.956      & 0.969         & 0.934         & 0.956 & 0.946  \\
        Ionosphere            & 0.822 & 0.813      & 0.915         & 0.858         & 0.804 & 0.732  \\
        Parkinson's           & 0.945 & 0.891      & 0.953         & 0.927         & 0.914 & 0.832  \\
        WDBC                  & 0.948 & 0.955      & 0.935         & 0.911         & 0.969 & 0.920  \\
        WPBC                  & 0.256 & 0.479      & 0.218         & 0.355         & 0.476 & 0.426  
        \end{tblr}
    }
\caption{Best CV score of tested algorithms on 10 datasets without feature reduction or selection.}\label{table1}
\end{table}

Considering the proposed algorithm performs well on datasets with lower dimensionality, sequential backward selection \cite{sbs} was implemented to see whether the model is able to produce more competitive results with reduced number of features. The results are provided in \ref{table2}.

\begin{table}[H]
    \centering
    \footnotesize
    \resizebox{\textwidth}{!}{ 
        \begin{tblr}{
            colspec={c|c|c|c|c|c|c},
            row{odd} = {gray!15}, % Light gray for odd rows
            row{even} = {white}   % White for even rows
        }
        Dataset               & KNN   & Perceptron & Random Forest & Decision Tree & SVM   & MPA    \\
        \hline
        Diabetes              & 0.811 & 0.759      & 0.824         & 0.746         & 0.824 & 0.792  \\
        Iris                  & 0.85  & 0.95       & 0.9           & 0.85          & 0.95  & 0.95   \\
        Penguin               & 1.0   & 1.0        & 1.0           & 1.0           & 1.0   & 1.0  \\
        Mammoraphic           & 0.865 & 0.839      & 0.849         & 0.823         & 0.865 & 0.844  \\
        SONAR                 & 0.952 & 0.904      & 0.904         & 0.952         & 0.880 & 0.880  \\
        Wisconsin(9 features) & 0.978 & 0.985      & 0.971         & 0.957         & 0.978 & 0.978  \\
        Ionosphere            & 0.985 & 0.943      & 0.957         & 0.957         & 0.929 & 0.915  \\
        Parkinson's           & 1.00  & 0.948      & 1.00          & 0.948         & 0.948 & 0.820  \\
        WDBC                  & 1.00  & 0.991      & 0.982         & 0.973         & 0.991 & 0.973  \\
        WPBC                  & 0.850 & 0.925      & 0.875         & 0.925         & 0.950 & 0.850  
        \end{tblr}
    }
\caption{Best scores after implementing sequential backward selection upto 3 features.}\label{table2}
\end{table}

Since a heuristic algorithm is a largely unexplored domain in machine learning, we should first establish whether the algorithm is an intelligent learner or not. Learning can be referred to as the process of correcting errors in the context of machine learning. 

\begin{figure}[H]
    \centering
    \begin{subfigure}[H]{0.45\textwidth}
        \centering
        \includegraphics[width=\textwidth]{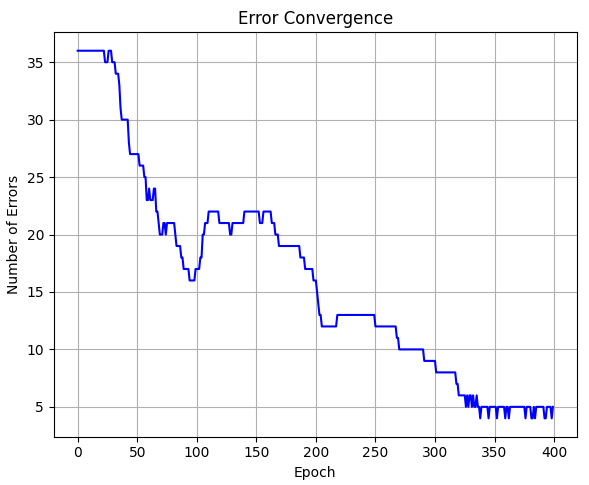}
        \caption{Convergence graph of MPA on Penguins dataset with two classes.}
        \label{fig:penguin_convergence}
    \end{subfigure}
    \hfill % Adds horizontal space between subfigures
    \begin{subfigure}[H]{0.45\textwidth}
        \centering
        \includegraphics[width=\textwidth]{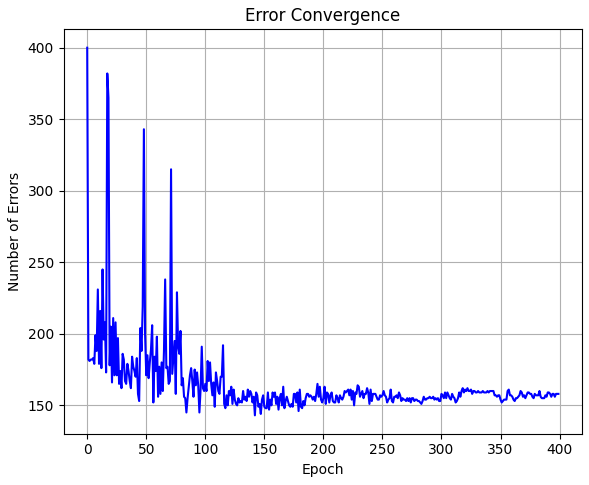}
        \caption{Convergence graph of MPA on Diabetes dataset with two classes.}
        \label{fig:diabetes_convergence}
    \end{subfigure}
    \caption{Convergence graphs of MPA on different datasets, illustrating varying learning behaviors.}
    \label{fig:convergence_comparison}
\end{figure}

Hence, whether the model learns or not can be shown by plotting a graph of errors vs. the progression of learning mechanism. A sample plot of convergence of the MPA is shown in \ref{fig:penguin_convergence} and \ref{fig:diabetes_convergence}.

The spikes correspond to sudden changes in the orientation of the hyperplane, which is often corrected in a few epochs but nevertheless indicates instability in the learning process. 

\subsection{Refining the algorithm}
It is important to note that in the previous methods described to initialize the model, emphasis was given to "smart initialization techniques" that initialized the decision boundary using the centroids of the two classes and derive a hyperplane perpendicular to the hyperplane on which both the centroids lie. Often, resulting in a highly optimized decision boundary or in local traps which would cause the errors to either remain constant or increase, i.e., diverge from the most optimum boundary. To address this, the initialization method was changed to completely random initialization. 

Although this helped show convergence graphs, the final resulting boundaries were not at par with contemporary algorithms. 
The algorithm was further modified to incorporate different stabilization techniques, which improved the performance of the model on the datasets tested. 

The results of this experiment are presented in \ref{table3}

\begin{table}[H]
    \centering
    \footnotesize
    \resizebox{\textwidth}{!}{ 
        \begin{tblr}{
            colspec={c|c|c|c|c|c|c},
            row{odd} = {gray!15}, % Light gray for odd rows
            row{even} = {white}   % White for even rows
        }
        Algorithm   & Dataset    & Best CV\footnote{cross validation} Score & Accuracy & Precision & Recall & F1 \\
        \hline
        ImprovedMPA & Iris       & 0.975 & 0.9   & 0.889 & 0.889 & 0.889 \\
        SVM-linear  & Iris       & 0.963 & 0.9   & 0.818 & 1.0   & 0.9   \\
        ImprovedMPA & Penguin    & 1.0   & 0.977 & 1.0   & 0.966 & 0.983 \\
        SVM-linear  & Penguin    & 0.994 & 1.0   & 1.0   & 1.0   & 1.0   \\
        ImprovedMPA & Parkinson  & 0.846 & 0.744 & 0.867 & 0.813 & 0.839 \\
        SVM-linear  & Parkinson  & 0.859 & 0.744 & 0.923 & 0.75  & 0.828 \\
        ImprovedMPA & Ionosphere & 0.850 & 0.803 & 1.0   & 0.391 & 0.563 \\
        SVM-linear  & Ionosphere & 0.868 & 0.887 & 0.895 & 0.739 & 0.810 \\
        ImprovedMPA & Parkinson  & 0.853 & 0.751 & 0.663 & 0.827 & 0.736 \\
        SVM-linear  & Parkinson  & 0.849 & 0.793 & 0.720 & 0.827 & 0.770 \\
        ImprovedMPA & Diabetes   & 0.759 & 0.746 & 0.682 & 0.545 & 0.606 \\
        SVM-linear  & Diabetes   & 0.764 & 0.812 & 0.810 & 0.618 & 0.701 \\
        ImprovedMPA & WDBC   & 0.87 & 0.92 & 0.93 & 0.95 & 0.94 \\
        SVM-linear  & WDBC   & 0.978 & 0.97 & 0.99 & 0.97 & 0.98 \\
        \end{tblr}
    }
\caption{Score after implementing proposed changes to the linear decision plane algorithm.}\label{table3}
\end{table}

It is important to note that these results were obtained after the algorithm trained with a randomized decision boundary with the points being constrained to be within the data ranges. 

The algorithm performs noticeably better when the data is scaled between the ranges of -100 to 100. Hence, this was taken as the default scaling values for all the test using sklearn's \cite{sklearn} MinMaxScaler. 

Theoretically, increasing the number of hyperplanes used for classification and training them on subsets of data could improve the algorithm by allowing it to generalize well.\footnote{Due to resource constraints, this hypothesis has not been tested at the time of writing of this paper.}

The changes were:
\begin{enumerate}
    \item Ensemble Approach: The classifier now trains multiple hyperplanes (default=3) and uses majority voting for prediction, improving robustness.
    \item Batch Processing: Updates are calculated in batches and applied together, leading to more stable convergence.
    \item Adaptive Learning Rate: More responsive learning rate adjustment with patience-based decay.
    \item Gradient Clipping: Prevents large, destabilizing updates by limiting step sizes based on data characteristics.
\end{enumerate}

\section{Usage of Splines in a Heuristics Based Algorithm}

After establishing that a heuristic algorithm can perform at par with SVM and other classifiers, we now move onto a new methodology used to create the decision boundary. Non linear decision boundaries are required because many real world datasets are non linear with complex relationships. In such cases, linear decision boundaries are not the best suited. Even with ensemble learning,
To compete with classification algorithms that create non linear decision boundaries, we will change the principle used to calculate the decision boundary from SVD to using cubic splines. Splines are widely used in numerical analysis and machine learning for their ability to create smooth, flexible curves that adapt to complex patterns in data. In this work, we employ cubic splines to define decision boundaries in SMPA. For a comprehensive introduction to cubic splines, see \cite{cubicsplines}.

For our purposes, we will use a Cubic spline and Piecewise Cubic Hermite Interpolating Polynomial splines\cite{pchip} from SciPy's \cite{scipy} scipy.interpolate module. 

\subsection{Reasons for experimenting with these splines}

We specifically chose these two methods for initial experimentation because:
\begin{enumerate}
    \item Cubic splines can have their derivatives at the end points to be set at zero. This helps stabilize the decision boundary and ensure the spline does not suffer from runge's phenomenon towards the end interpolating points see \cite{runge}.
    \item Cubic splines are naturally smooth and at the same good at approximating curves. This makes them an ideal candidate for a heuristic point based method.
    \item PCHIP are great for avoiding overshooting while maintaining smooth transitions. 
\end{enumerate}

Following the methodology used for creation of the MPA, we first create an algorithm for two dimensional dataset and build forward from it. This approach helps in visualization and getting an intuitive sense of how the boundary moves. 

\subsection{Adaptive Heuristic Updates}
The new algorithm was created by only changing the function used to calculate the spline. However, directly changing the type of decision boundary does not create a non-linear classifier because with a non-linear curve the challenge of stabilization arises which is difficult to achieve without introducing some additional mechanics. These improvements gave rise to Improved Spline Moving Points Algorithm - ImprovedSMPA. The improvements which help it achieve the results which will be discussed in the next section are:
\begin{itemize}
    \item Error-Based Adjustments: Misclassified points are identified by comparing their vertical displacement from the current spline boundary. An adaptive margin, influenced by the distance of a point from the nearest control point, scales the update step. This is controlled by a lambda scaling strategy (log, sqrt, or none), which intelligently adjusts the update magnitude based on local geometry.
    \item Learning Rate Scheduling: A decaying learning rate mechanism is employed, with the rate reduced when improvements stagnate over a preset patience period. This ensures a fine-tuning of the spline boundary as the algorithm converges, ultimately preventing overshooting.
    \item Control Point Update Dynamics: For each misclassified sample, the algorithm calculates a corrective step. When possible, it leverages the position of correctly classified opposite class points to compute a more informed directional update. In cases lacking such guidance, a conservative vertical adjustment is applied. Special care is taken for boundary control points to prevent erratic behavior at the edges.
\end{itemize}

Using these adaptive steps, the algorithm was tested on 2 dimensional moons and blobs datasets from sklearn's \cite{sklearn} dataset creation library. 

The testing methodology used to evaluate the Spline Moving Points Algorithm (SMPA) and baseline classifiers follows a structured approach incorporating repeated train-test splits, feature scaling, hyperparameter optimization, and statistical analysis.

\subsubsection{Methodology Used}
The testing methodology used to evaluate the Spline Moving Points Algorithm (SMPA) and baseline classifiers follows a structured approach incorporating repeated train-test splits, feature scaling, hyperparameter optimization, and statistical analysis.

\begin{enumerate}
    \item Dataset Generation and Preprocessing: 
    A synthetic dataset is generated using the make moons function and make blobs, introducing controlled noise to assess classifier robustness. Each classifier undergoes feature scaling with a dedicated scaler type to ensure compatibility with its underlying algorithm.

    \item Repeated Train-Test Splitting:
    Instead of a single static split, the dataset is randomly partitioned into training and test sets across multiple iterations (50 runs). This approach enhances robustness by evaluating model performance across different data distributions.

    \item Hyperparameter Optimization:
    A grid search is performed for each classifier using Stratified K-Fold cross-validation (3 folds). The grid search explores predefined hyperparameter spaces tailored to each classifier, ensuring optimal configurations for fair comparisons.

    \item Performance Evaluation and Visualization:
    Each classifier is evaluated on an independent test set, and accuracy scores are recorded. SMPA-specific visualizations, including decision boundary plots and convergence graphs, are generated to analyze its behavior.

    \item Stability Testing:
    The classification process is repeated over 5 and 25 independent runs, and statistical measures such as mean accuracy and standard deviation are computed to assess consistency and reliability.

    \item Statistical Significance Testing:
    The final evaluation involves a pairwise t-test between SMPA and baseline classifiers to determine whether performance differences are statistically significant. A significance threshold of p $> 0.05$ is used to identify meaningful improvements.

\end{enumerate}

This methodology ensures a rigorous, reproducible comparison between SMPA and standard classifiers, accounting for variability in training data, hyperparameter selection, and classifier stability.

\subsubsection{Results}

Results on moons dataset for 5 runs is given in Table \ref{metrics5run}

\begin{table}[H]
    \centering
    \footnotesize
    \resizebox{\textwidth}{!}{ 
        \begin{tblr}{
            colspec={c|c|c|c|c},
            row{odd} = {gray!15}, % Light gray for odd rows
            row{even} = {white}   % White for even rows
        }
        Classifier & Mean Accuracy & Std Deviation & T-Statistic (compared to SMPA) & P-Value (compared to SMPA) \\
        \hline
        SMPA       & 0.9825        & 0.01          & NA                             & NA                         \\
        SVM        & 0.9725        & 0.0094        & 1.4606                         & 0.1823                     \\
        RF         & 0.9725        & 0.0094        & 1.4606                         & 0.1823                     \\
        DT         & 0.9700        & 0.015         & 1.3868                         & 0.2029                    
        \end{tblr}
    }
\caption{Metrics of SMPA on 5 runs based on the testing methodology.}\label{metrics5run}
\end{table}

As can be observed from the data, there is no statistically significant difference between the classifiers and SMPA which strengthens its position as a classification algorithm. It manages to perform at par with other classification algorithms and the mean scores. However, the lack of statistical significance in t-tests (p $>$ 0.05) suggests that the observed performance differences could be due to chance rather than a fundamental advantage.

Extending the test to include more runs, let us look at the results from 25 runs using the same methodology. Results are given in Table \ref{metrics25run}

\begin{table}[H]
    \centering
    \footnotesize
    \resizebox{\textwidth}{!}{ 
        \begin{tblr}{
            colspec={c|c|c|c|c},
            row{odd} = {gray!15}, % Light gray for odd rows
            row{even} = {white}   % White for even rows
        }
        Classifier & Mean Accuracy & Std Deviation & T-Statistic (compared to SMPA) & P-Value (compared to SMPA) \\
        \hline
        SMPA       & 0.9690        & 0.0150          & NA                             & NA                         \\
        SVM        & 0.9660        & 0.0192          & 0.6023                         & 0.5498                     \\
        RF         & 0.9675       & 0.0177        & 0.3165                         & 0.7530                     \\
        DT         & 0.9615        & 0.0197         & 1.4840                         & 0.1443                    
        \end{tblr}
    }
\caption{Metrics of SMPA on 25 runs based on the testing methodology.}\label{metrics25run}
\end{table}

Even with 25 runs, SMPA has maintained highest mean accuracy and also displayed the lowest standard deviation between results. But, still no statistically significant difference can be established. 

Moving onto the blobs dataset, we see similar results in Table \ref{blobs_5run}

\begin{table}[H]
    \centering
    \footnotesize
    \resizebox{\textwidth}{!}{ 
        \begin{tblr}{
            colspec={c|c|c|c|c},
            row{odd} = {gray!15}, % Light gray for odd rows
            row{even} = {white}   % White for even rows
        }
        Classifier & Mean Accuracy & Std Deviation & T-Statistic (compared to SMPA) & P-Value (compared to SMPA) \\
        \hline
        SMPA       & 0.990        & 0.009          & NA                             & NA                         \\
        SVM        & 0.988        & 0.008          & 0.408                          & 0.6938                     \\
        RF         & 0.985        & 0.009          & 0.756                          & 1.314                    \\
        DT         & 0.983        & 0.917          & 1.342                          & 0.2165                    
        \end{tblr}
    }
\caption{Metrics of SMPA on 5 runs on blobs based on the testing methodology.}\label{blobs_5run}
\end{table}

SMPA performs marginally better than SVM and RF but without statistically significant improvements (p $>$ 0.05). The Decision Tree shows the weakest performance and highest variability. While SMPA appears robust, the results suggest that the differences in accuracy may not be practically significant. Further testing on more complex datasets is necessary to assess its true advantage.

\subsection{Conclusion}
This study introduced and evaluated the Spline Moving Points Algorithm (SMPA) as an extension of the Moving Points Algorithm (MPA), incorporating cubic splines to enable non-linear decision boundaries. The experimental results demonstrated that SMPA achieves accuracy comparable to established classifiers, such as Support Vector Machines (SVM), Random Forest (RF), and Decision Trees (DT), across multiple synthetic datasets.

\textbf{Key findings include:}
\begin{enumerate}
    \item \textbf{SMPA consistently achieved the highest mean accuracy} across tested datasets, with lower variance in results.
    \item Despite superior accuracy, \textbf{statistical significance tests (t-tests) showed no meaningful difference} between SMPA and other classifiers (p $>$ 0.05), suggesting that performance gains may not be substantial in practical settings.
    \item The introduction of \textbf{splines improved decision boundary flexibility}, but additional refinements, such as stability mechanisms and adaptive updates, were necessary to ensure convergence and prevent overfitting.
\end{enumerate}

While SMPA shows promise as an alternative classification method, particularly for non-linear datasets, further evaluation on real-world, high-dimensional datasets is required to determine its broader applicability. Future work will explore scalability, computational efficiency, and generalization to more complex problems.

\end{document}